# Analysis of Voluntarily Reported Data Post Mesh Implantation for Detecting Public Emotion and Identifying Concern Reports

Indu Bala[1], Lewis Mitchell[1], Marianne H Gillam[2]

[1] The University of Adelaide, Adelaide, 5005, Australia
[2] University of South Australia, Adelaide, 5001, Australia
iinduyadav@gmail.com; indu.bala@adelaide.edu.au

**Abstract.** Mesh implants are widely utilized in hernia repair surgeries, but post-operative complications present a significant concern. This study analyses patient reports from the Manufacturer and User Facility Device Experience (MAUDE) database spanning 2000 to 2021 to investigate the emotional aspects of patients following mesh implantation using Natural Language Processing (NLP). Employing the National Research Council Canada (NRC) Emotion Lexicon and TextBlob for sentiment analysis, the research categorizes patient narratives into eight emotions—anger, fear, anticipation, trust, surprise, sadness, joy, and disgust—and assesses sentiment polarity. The goal is to discern patterns in patient sentiment over time and to identify reports signaling urgent concerns, referred to as 'Concern Reports,' thereby understanding shifts in patient experiences in relation to changes in medical device regulation and technological advancements in healthcare. The study detected an uptick in Concern Reports and heightened public emotion during the periods of 2011-2012 and 2017-2018. Through temporal analysis of Concern Reports and overall sentiment, this research provides valuable insights for healthcare practitioners, enhancing their understanding of patient experiences post-surgery, which is critical for improving preoperative counselling, postoperative care, and preparing patients for mesh implant surgeries. The study underscores the importance of emotional considerations in medical practices and the potential for sentiment analysis to inform and enhance patient care.

**Keywords:** Mesh implant surgery, Sentiment analysis, Emotion extraction, Manufacturer and User Facility Device Experience (MAUDE) database, Post-operative complications, Concern Reports.

## 1 Introduction

The convergence of healthcare and technology has continuously evolved in patient care over time. Among the various techniques aiding medical decision-making, opinion mining or sentiment analysis stands out as pivotal. This involves the application of Natural Language Processing (NLP) techniques to analyze unstructured textual data, a practice increasingly embraced across diverse sectors, including healthcare. Through NLP, significant insights can be gleaned from vast amounts of unstructured text, unveiling patterns, sentiments, and emotional cues that offer a profound understanding of patient experiences [1].

In recent years, the significance of sentiment analysis in capturing people's emotions and bolstering decision-making, especially in the healthcare sector, has been increasingly highlighted. For instance, Mondel et al. [2] introduced an innovative domain-specific lexicon combined with machine learning techniques to identify semantic relationships. Their approach systematically categorized medical concepts such as diseases, symptoms, drugs, and human anatomy, evaluating sentiments as either positive or negative. In a similar vein, Ramírez-Tinoco et al. [3] developed a sentiment analysis module to extract sentiments and emotions at both the comment and entity levels in healthcare texts. Extending the application of sentiment analysis, Chintalapudi et al. [4] examined seafarers' medical documents, Alam et al. [5] analyzed responses to COVID-19 vaccination on Twitter, Hu [6] investigated texts related to public health emergencies, and Edara et al. [7] explored cancer patients' experiences. These studies collectively demonstrate the utility of sentiment analysis in providing invaluable insights for medical professionals, aiding in informed decision-making.

Yuvaraj et al. [8] investigated the classification of electroencephalogram (EEG) emotional states, focusing on valence and arousal. They examined the classification accuracy of five EEG feature sets, those derived using wavelet analysis. Their feature-classification method achieved good classification accuracy for valence across the datasets used. Similarly, Wang et al. [9] developed a method to simultaneously identify emotions and their corresponding causes within conversations spanning various modalities. To evaluate this method, they established two benchmark systems: one based on a heuristic approach that analyzes the inherent patterns in the positioning of causes and emotions, and another that utilizes a deep learning strategy to integrate multimodal features for the extraction of emotion-cause pairs. They also conducted a human performance test for comparison. Denecke et al. [10] provided a comprehensive review emphasizing the role of sentiment and emotion analysis in clinical notes, noting that such text is a rich source of information that significantly aids decision-making processes.

Our study also utilizes this rich free-text from the reports of hernia mesh implant surgery submitted to the Manufacturer and User Facility Device Experience (MAUDE) database, which is regulated by the US Food and Drug Administration (FDA) [11-12]. The detailed information regarding the data and mesh implant will be further discussed in the next section of the literature review. This chapter employs NLP methodologies, including sentiment identification using the TextBlob library and emotion detection with the National Research Council Canada (NRC) Emotion Lexicon, to process and interpret the textual content of the reports. The analysis aims to uncover patterns and shifts in patient experiences, providing insights to guide future enhancements in device design, patient care, and regulatory measures. The study's objectives include:

- Determining the sentiment in patient reports and extracting emotions using the 8 core emotions framework to better understand their experiences and feelings.

- Identifying reports that signal significant concerns within the dataset, indicating areas needing urgent attention.
- Analyzing year-by-year temporal trends in emotional expressions and reported concerns, assisting healthcare professionals in making informed decisions.

The organization of the study is as follows: Section 2 provides a detail literature review of mesh implantation, Section 3 presents the methodologies employed, Section 4 discusses the experimental settings and results, and Section 5 concludes the findings and outlines directions for future research.

## 2 Literature Review

Mesh is a versatile material primarily used in medical applications, especially in surgical procedures. It typically consists of synthetic materials such as polypropylene and polyester, or biological compounds, and serves to support and reinforce weakened or damaged tissues. In the realm of medicine, mesh is most frequently used in hernia repair surgeries to stabilize the affected area and prevent hernia recurrence by reinforcing the abdominal wall [12]. Mesh is highly valued for its durability and effectiveness in providing long-term reinforcement across various surgical repairs. Despite its benefits, mesh implants can lead to significant postoperative complications, including pain, infection, and stress, which can adversely affect a patient's quality of life [12-13].

The use of mesh implants has long been a subject of concern and controversy. For instance, between 2005 and 2007, the FDA recorded over a thousand adverse events related to the use of mesh in hernia repair procedures [14]. In response, the FDA issued a Public Health Notification in 2008 to alert healthcare providers and patients about serious complications associated with these devices. The media further highlighted the issue, with reports on lawsuits and case studies, notably those discussed by Voreacos and Nussbaum in 2011 [15]. More recent scrutiny in the UK and New Zealand has focused on the repercussions of companies withdrawing mesh products from the market, emphasizing the ongoing need for rigorous oversight of mesh implant surgeries and a comprehensive understanding of patient post-operative experiences [16-18]. Due to global concerns regarding mesh implants, many international discussions have taken place, and new regulations concerning mesh implants have been developed [19].

Due to these controversies, extensive statistical research has been conducted to improve the quality of mesh materials, trends in materials, manufacturing processes, therapeutic approaches, and regulations concerning mesh implants. Mangir et al. [13] discussed the complications associated with mesh implants and their severity in clinical practices through case studies. Farmer et al. [21] explored the use of thermoplastic polyurethane produced by fused deposition modeling as an alternative mesh material, proposing it as a potential avenue for creating a new generation of safer mesh implants.

Despite extensive research on the physical impacts of mesh implants, there is a significant lack of studies addressing the intersection of emotional analysis and patient-reported outcomes in this area. This research aims to fill this gap by providing detailed insights into the emotional narratives of patients, which have often been overlooked in traditional clinical research on postoperative complications from mesh implants. The findings of this study could lead to significant improvements in patient care strategies and contribute to more holistic healthcare practices.

To support this narrative, we utilize the MAUDE database, which serves as a comprehensive repository of narratives related to implantable medical devices [21]. These narratives fulfill regulatory requirements and offer invaluable insights into patient experiences, device malfunctions, and adverse events. They reveal the emotional and psychological impacts of implanted medical devices, aspects that are frequently ignored. The objective of this chapter is to explore the experiences of patients after mesh implantation surgery as reported in the MAUDE database and to identify the most concerning reports.

# 3 Methods

## 3.1 Data Collection and preprocessing

The dataset, obtained from the Food and Drug Administration, is an open repository that houses record critical to medical device safety [11]. It consists of four primary file types—Master Event, Device, Patient, Text—and two supplementary file types—Device Problems and Problem Code Descriptions. Together, these files offer a detailed overview of adverse events or product problem reports [21]. For our study, we specifically focused on data related to mesh medical devices extracted from these complex and voluminous files, with a particular interest in reports categorized as voluntary submissions. Covering the period from 2000 to 2021, our analysis included 2,422 reports across 125 columns, primarily leveraging columns that contain voluntarily submitted free-text entries, alongside their reported dates and years.

To ensure the quality of the data, reduce noise, and improve coherence, we implemented a series of preprocessing measures utilizing Python. Initially, preprocessing involved removing duplicate rows, converting text to lowercase, and eliminating numerical values, punctuation, and irrelevant terms [22]. We then employed the NLTK package for standard tokenization, lemmatization, part-of-speech tagging, and stop-word removal from the Patient Outcome Reports (PORs) [23].

## 3.2 Sentiment Analysis

In the methodological framework of our study, we address the complexities of analyzing unstructured medical texts, a challenge intensified by the specialized language prevalent in the medical domain. To navigate the inherent difficulty of accurately capturing sentiments expressed in such specialized language, our methodology employs a dual approach focusing on sentiment and emotion analysis, utilizing TextBlob in conjunction with the NRC Emotion Lexicon.

The initial phase of our analysis utilizes TextBlob, a versatile tool within the NLP field [24]. TextBlob offers an accessible API that integrates with foundational NLP libraries like NLTK and Pattern, streamlining a variety of NLP tasks including sentiment analysis. This integration enables TextBlob to facilitate diverse linguistic analyses, ranging from part-of-speech tagging to detailed sentiment assessments. Central to TextBlob sentiment analysis is the derivation of polarity scores, quantifying the emotional valence of text on a scale from -1.0 (entirely negative) to 1.0 (entirely positive), alongside subjectivity scores that range from 0.0 (fully objective) to 1.0 (fully subjective). This process aggregates individual word sentiment values into an overall sentiment score, succinctly capturing the text's emotional essence [6, 25].

To systematically categorize sentiments derived from patient reports, we defined specific thresholds: texts with polarity scores equal to or greater than 0.05 are classified as positive; those with scores between -0.05 and 0.05 are deemed neutral; and texts scoring less than or equal to -0.05 are considered negative. This classification is essential for parsing the array of sentiments within patient narratives [25].

Our analysis is further enhanced by incorporating the NRC Emotion Lexicon [26], a rich collection of English words associated with eight primary emotions: anger, fear, anticipation, trust, surprise, sadness, joy, and disgust. The association of words with specific emotions aids in constructing a detailed emotional profile for the analyzed texts [27, 28]. This dual analytical approach not only enriches the sentiment analysis performed with TextBlob but also presents a comprehensive view of the emotional dynamics in patient reports.

### 3.3 Identify the Concerns Reports

In our study, we further delve deeper into patient narratives to distinguish reports that exhibit a significant degree of severity, which we term as ‘Concern Reports’. To systematically identify these reports, we adopted a multi-faceted analytical approach, focusing on sentence-level analysis to ensure precision. Our methodology revolves around three key metrics: the negativity ratio, the mean negative score, and the mean polarity score, each offering a distinct perspective on the sentiment embedded in the reports. We define these statistics as follows:

**Case 1: Negative sentence ($\mathbf{S_{neg}}$):** A sentence $S$ is defined as negative, $S_{neg}$, if NRC lexicon detects emotion $E \in \{fear, anger, sadness, disgust\}$ with intensity ≥ 0.05 or polarity score $P_s \leq$ -0.05. Mathematically it can be written as

$$S_{neg} = \begin{Bmatrix} 1, & if\ E \geq 0.05\ or\ P_s \leq -0.05 \\ 0, & otherwise \end{Bmatrix} \quad (1)$$

**Case 2: Negativity ratio ($\mathbf{R_{neg}}$):** The negativity ratio denoted as $R_{neg}$, of a report is defined as the ratio of the number of negative sentences $S_{neg}$ to the total number of sentences $S_{total}$ in the report such as : $R_{neg} = \frac{n(S_{neg})}{S_{total}}$ (2)

**Case 3: Mean negative score ($\mathbf{A_{neg}}$):** Let $S_{neg_score_i}$ denote the negative score of the $ith$ sentence identified as negative, where the score reflects the intensity of negativity (based on emotion intensity or negative polarity). If a report contains n sentences identified as negative, the Mean Negative Score ( $A_{neg}$) of the report is defined as the sum of all negative scores divided by the number of such sentences (say $n$): $A_{neg} = \frac{\sum_{i=1}^{n} S_{neg_score_i}}{n}$ (3)

**Case 4: Mean polarity score:** It is denoted by $A_{pol}$, is the mean of all TextBlob polarity score $P_s$ of the sentences in the reports: $A_{pol} = \frac{\sum P_s}{S_{total}}$ (4)

**Case 5: Concern Reports:** In the context of patient reports after mesh implantation for hernia surgery, a report is evaluated for generating concerns, which could include significant negativity or indications of complications, based on its negativity ratio ($R_{neg}$), mean negative score ($A_{neg}$), and an mean polarity score ($A_{pol}$), utilizing predefined thresholds ( $\delta_1, \delta_2, \delta_3$). These thresholds are determined empirically and range from 0.0 to 0.5.

A report is flagged for generating concerns if the following conditions are met:

$$\begin{Bmatrix} 1, \ if \ R_{neg} > \delta_1 \ and \ \left(A_{neg} > \delta_2 \ ir \ A_{pol} > \delta_3\right) \\ 0, \qquad\qquad\qquad\qquad otherwise \end{Bmatrix} \quad (5)$$

Case 5, employs a structured approach to meticulously shift through patient reports, identifying those that not only present a high severity level—as indicated by the negativity ratio ($R_{neg}$)—but also illuminate particular areas of significant concern. This is achieved either through elevated mean negative scores ($A_{neg}$) or notable overall sentiment biases ($A_{pol}$). Specifically, a report crossing the threshold value ($\delta_1$) for $R_{neg}$ is indicative of a prevailing negative sentiment. Further scrutiny is applied if such a report also exhibits a mean negative score surpassing the threshold ($\delta_2$), signifying pronounced negative experiences, or a mean polarity score exceeding ($\delta_3$), denoting a distinct negative sentiment bias. Such reports are flagged as expressing severe concerns, warranting immediate attention or in-depth investigation. This discerning identification process is underpinned by a meticulously calibrated set of threshold values $\delta_1$=0.35, $\delta_2$=0.4, and $\delta_3$=0.4. These were not arbitrarily selected but rather derived from a comprehensive and careful examination of the reports, further refined by an extensive data fine-tuning process [25]. Our approach prioritizes an in-depth understanding of the data's distribution, ensuring our analysis is firmly rooted in the empirical evidence presented by the reports. By doing so, we ensure that the threshold values optimally reflect the nuanced variations and patterns inherent in the patient narratives. Ultimately, this methodological rigor enables the isolation of reports that are not merely quantitatively negative but qualitatively severe in their expression of patient concerns and experiences.

# 4 Experimental Results and Discussion

## 4.1 Measuring Sentiments and Emotions

This section presents the findings from our analysis of 2,422 patient reports, which include 32,476 sentences. Using TextBlob, we initially classified the sentiments within these reports into positive, negative, and neutral categories. We then delved deeper to extract eight specific emotions with the help of the NRC Emotion Lexicon, using Python and its libraries for an in-depth data analysis.

Our study highlights a significant trend of negative sentiment towards mesh implant devices, with 2,048 of the analyzed reports indicating adverse experiences. In contrast, 270 reports were found to be neutral, and 104 were positive. These results, spanning from 2000 to 2021, are compiled in Table 1, which offers a comprehensive overview of the sentiment distribution and the mean polarity scores for each sentiment category, underscoring the predominance of negative feedback in patient narratives.

The range of mean polarity scores, from -0.146 to -0.725, underscores the overarching negative perception of mesh implant devices among patients. The annual fluctuations in sentiment are highlighted by years such as 2011 and 2012, which exhibited a notable spike in negative reports.

Complementing the sentiment analysis, the extraction of emotions using the NRC lexicon revealed a comprehensive picture of patients' emotional responses, as depicted in Fig.1. The graph illustrates the prevalence of eight key emotions from 2000 to 2021—ranging from fear and anger to joy and trust—across patient narratives. There is a clear upward trend in the expression of all sentiments, indicating an increased level of emotional reporting in patient narratives over time. Notably, negative emotions such as fear, anger, and sadness dominate the chart, with fear being the most reported sentiment in the latter years. This trend suggests growing concerns about complications from mesh implants.

Sentiments of disgust and anticipation also show an upward trend, albeit less pronounced than the primary negative emotions. Interestingly, positive sentiments such as joy and trust are present but are significantly lower in proportion and remain relatively stable over time, suggesting that positive experiences or outcomes are less frequently reported in the context of these surgeries. This may reflect a decline in patient satisfaction or confidence in mesh implants. The increasing mention of surprise over time may indicate patients' unexpected outcomes post-surgery.

These emotional trajectories, consistent with trends observed in the sentiment analysis, demonstrate notable surges, particularly in the years 2011-2012 and again in 2017-2018, which align with findings from previous research [12]. The scarcity of neutral sentiments implies that patients typically have definitive opinions about their experiences with mesh implants, with very few remaining ambivalent. The low occurrence of positive sentiments highlights a tendency among patients to report negative outcomes more frequently than positive ones.

In conclusion, our findings underscore the importance for medical practitioners to be attentive to these reported sentiments and emotions, as they provide critical insights into patient experiences following mesh implant surgery. Recognizing periods of heightened patient dissatisfaction, such as those in 2011-2012, is crucial for identifying the root causes of negative experiences and for enhancing patient care and outcomes.

### 4.2 Analysis of Concern reports

In our analysis of 2,422 reports related to mesh implant surgeries, we concentrated on the 2,048 reports that conveyed negative sentiments and set aside the positive and neutral ones. Within these, we distinguished a significant portion—about 41%—as 'concern reports', marking them for special attention due to their serious nature. Fig. 2 illustrates the year-wise distribution of these concern reports. Despite some variability, there is a clear trend consistent with the overall number of reports and those expressing negative sentiments. This trend suggests that the incidence of severe complications correlates with the volume of surgeries performed. The graph notably displays periodic spikes in concern reports, indicating times of heightened complications. Such patterns could signal issues with surgical techniques, mesh types, or patient selection criteria during those times. For instance, a surge in concern reports around the years 2011-2012 prompts a critical evaluation of related factors, as also underscored by previous observations in the study. Complementing this, our regression analysis—illustrated in Fig. 3—analyzes the relationship between the total number of surgeries and the concern reports, uncovering a positive correlation. The implication is significant: as surgical frequency goes up, so does the incidence of concern reports, suggesting a connection between surgical volume and complications. This correlation transcends statistical significance, pointing towards real-world implications for surgical practice. Outliers in the regression graph, notably above the predictive line, are indicative of exceptional periods of concern. These anomalies, particularly pronounced during the years 2011-2012, warrant further scrutiny to determine if external influences such as changes in surgical practices, mesh types, or broader healthcare contexts may have contributed to these discrepancies.

**Table 1.** Yearly Distribution of Patient Sentiments and Mean Polarity Scores

| Year | Total Reports | Negative (%) | Positive (%) | Neutral (%) | Mean Polarity Score |
|---|---|---|---|---|---|
| 2000 | 6 | 100 | 0 | 0 | -0.304117 |
| 2001 | 9 | 77 | 0 | 23 | -0.6344 |
| 2002 | 6 | 83 | 17 | 0 | -0.511267 |
| 2003 | 25 | 100 | 0 | 0 | -0.222652 |
| 2004 | 18 | 50 | 17 | 33 | -0.374194 |
| 2005 | 13 | 100 | 0 | 0 | -0.146646 |
| 2006 | 19 | 57 | 36 | 7 | -0.120232 |
| 2007 | 88 | 86 | 4 | 10 | -0.683107 |
| 2008 | 79 | 88 | 3 | 8 | -0.616648 |
| 2009 | 86 | 83 | 9 | 8 | -0.644405 |
| 2010 | 56 | 89 | 2 | 9 | -0.725543 |
| 2011 | 287 | 86 | 3 | 11 | -0.712722 |
| 2012 | 249 | 81 | 5 | 13 | -0.683901 |
| 2013 | 212 | 85 | 5 | 8 | -0.538248 |
| 2014 | 189 | 86 | 1 | 12 | -0.67151 |
| 2015 | 200 | 82 | 6 | 12 | -0.63968 |
| 2016 | 168 | 92 | 2 | 6 | -0.625582 |
| 2017 | 231 | 86 | 3 | 11 | -0.569468 |
| 2018 | 152 | 82 | 4 | 14 | -0.628941 |
| 2019 | 190 | 85 | 4 | 11 | -0.333576 |
| 2020 | 70 | 81 | 12 | 7 | -0.496041 |
| 2021 | 69 | 76 | 9 | 15 | -0.352309 |

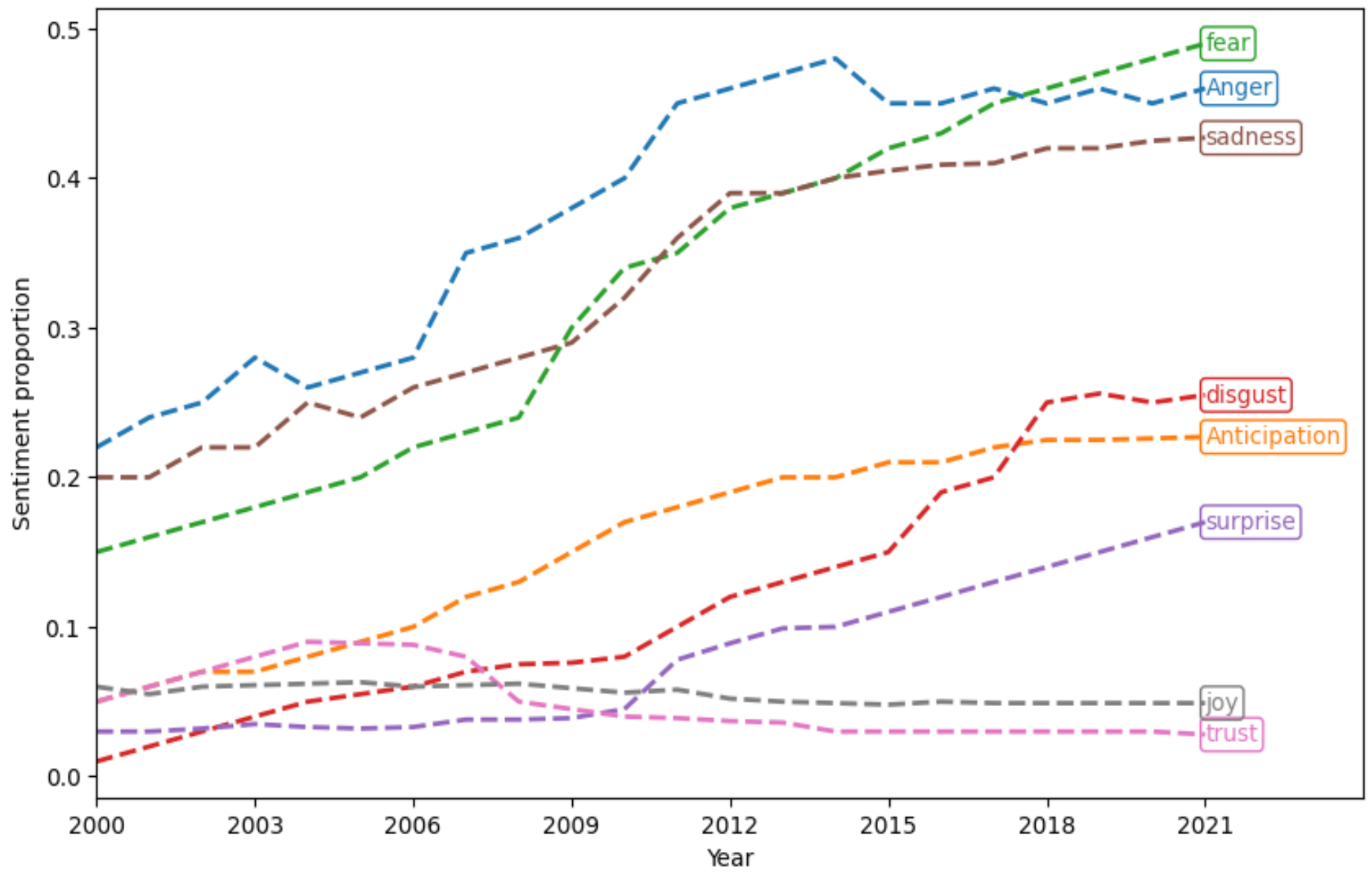

**Fig. 1.** Yearwise Emotion Evolution in report post-Mesh Implant surgery

In Fig. 4, a word cloud derived from the concern reports serves as a stark visual representation of patient distress. Dominant terms like "pain," "helpless," "fear," and "mesh" paint a telling picture of the patient's experience, with "pain" and "helpless" notably prominent, underscoring the physical and emotional burden of the patients. "fear" and "scared" imply apprehension about the surgery or its aftermath, possibly reflecting worries about complications or the effectiveness of the implants. Words like "severe," "misery," "torment," and "unbearable" highlight the intensity of negative experiences. The presence of words related to complications ("infection," "erosion," "obstruction") further corroborates the data, which showed an uptick in concern during certain years. These insights align with the negative sentiments revealed in the analysis, where a high percentage of reports were categorized as negative, supported by low mean polarity scores.

Together, these analyses emphasize the critical need for ongoing monitoring and proactive intervention in mesh implant surgeries. By bringing these findings to light, our research aims to facilitate enhancements in surgical procedures, patient care, and regulatory oversight to improve patient outcomes and uphold the highest standards of healthcare.

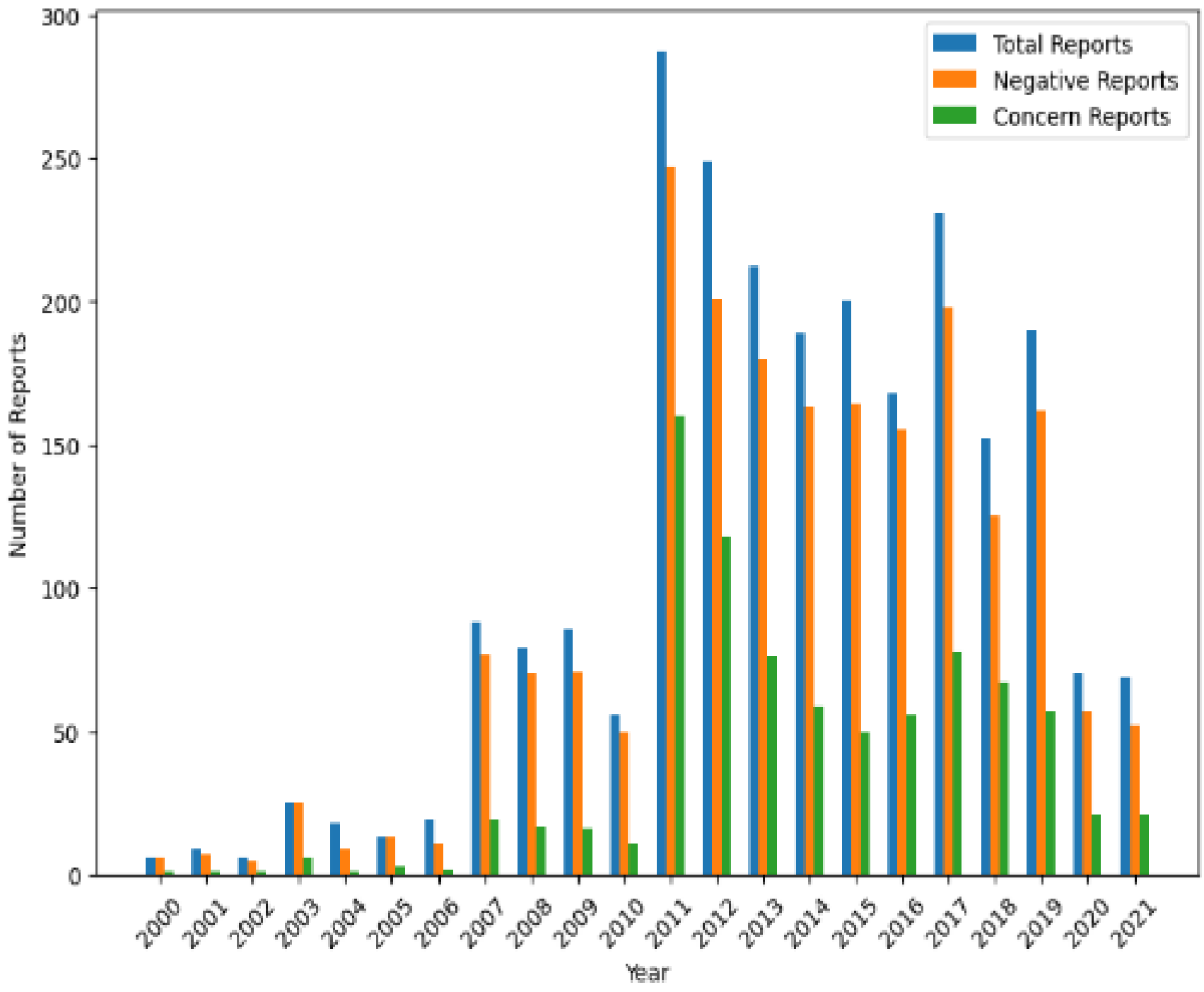

**Fig. 2.** Yearly trends of reports

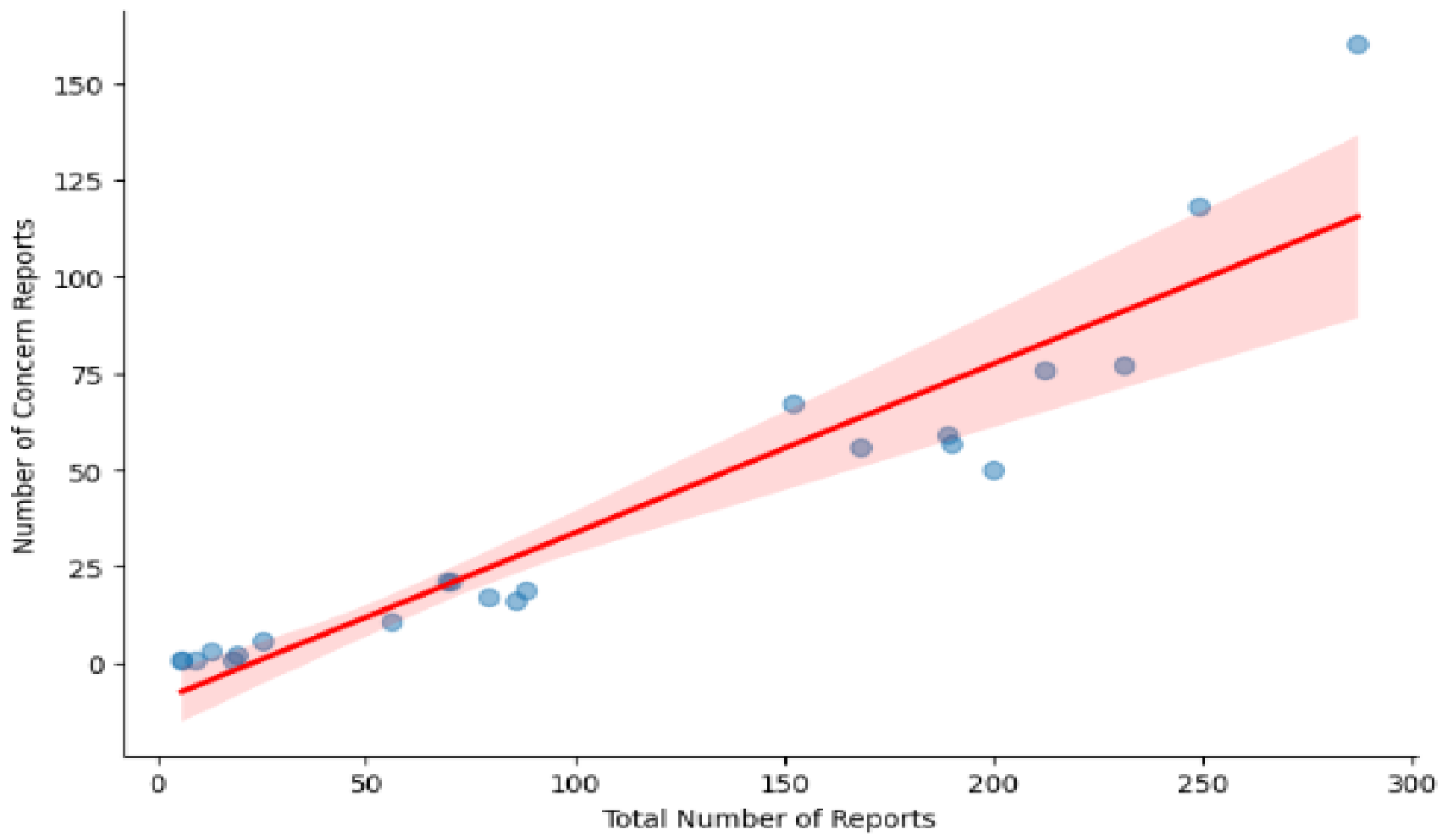

**Fig. 3.** Regression analysis of Concerns Reports.

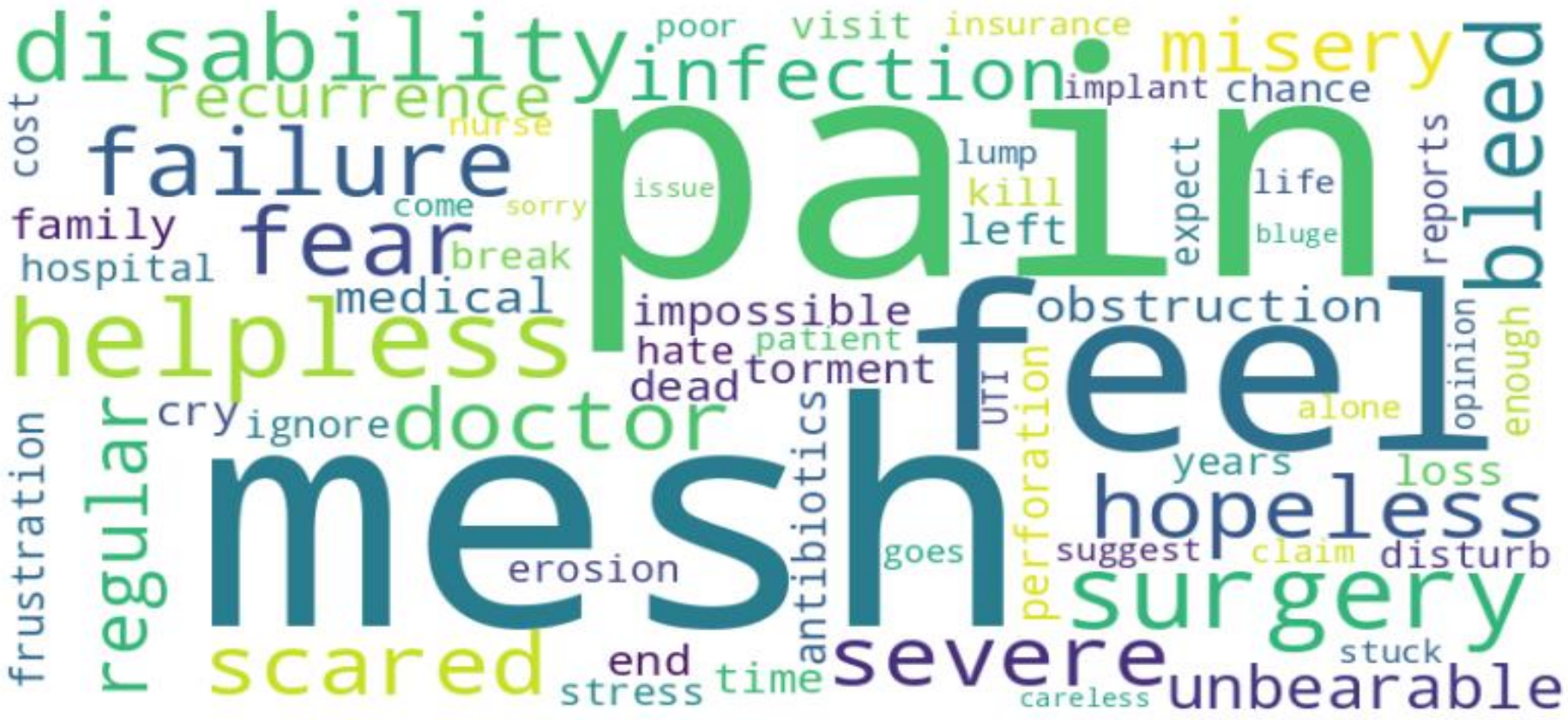

**Fig. 4.** Word Cloud of Concern Reports.

## 5 Conclusion

This study presents a comprehensive sentiment analysis of patient reports related to mesh implant surgeries, examining narratives from the MAUDE database spanning 2000 to 2021. We utilized Natural Language Processing tools, specifically the NRC Emotion Lexicon and TextBlob, to decode a wide range of emotions expressed by patients' post-surgery, including anger, fear, anticipation, trust, surprise, sadness, joy, and disgust. We also developed an algorithm that helps identify concern reports from the negative set of feedback. These concern reports aim to highlight areas that require prompt attention. Our investigation revealed that a significant portion of the negative reports—especially during the periods 2011-2012 and 2017-2018—showed a surge in concern reports as well as negative emotions, suggesting systemic issues with surgical practices or the mesh devices themselves. Our study strategically demonstrated the increase in concern reporting during these periods using regression analysis. The data show a pronounced prevalence of negative sentiments within the patient feedback, particularly emotions associated with pain, fear, and helplessness. These findings underscore the profound emotional and physical impact of complications arising from mesh implants. This emphasizes the crucial need to consider patient emotions in medical practice to better prepare and support patients throughout their surgical experience.

Our study not only provides critical insights into the emotional outcomes of surgical interventions but also highlights the potential of sentiment analysis as a tool for enhancing patient care. In future, the application of more sophisticated analytical techniques and broader datasets, including medical records and clinical outcomes, could further this research. The incorporation of heuristic algorithms to refine the thresholds for Concern Reports might yield more nuanced understanding, ultimately contributing to the elevation of patient safety and satisfaction in the surgical field.

## References

1. Bala, I. (2023). Natural Language Processing in Medical Science and Healthcare. Medicon Medical Sciences, 4(1), 1-2.
2. Mondal, A., Cambria, E., Das, D., Hussain, A., & Bandyopadhyay, S. (2018). Relation extraction of medical concepts using categorization and sentiment analysis. Cognitive Computation, 10, 670-685.
3. Ramírez-Tinoco, F. J., Alor-Hernández, G., Sánchez-Cervantes, J. L., Salas-Zárate, M. D. P., & Valencia-García, R. (2019). Use of sentiment analysis techniques in healthcare domain. Current trends in semantic web technologies: theory and practice, 189-212.
4. Chintalapudi, N., Battineni, G., Di Canio, M., Sagaro, G. G., & Amenta, F. (2021). Text mining with sentiment analysis on seafarers' medical documents. International Journal of Information Management Data Insights, 1(1), 100005.
5. Alam, K. N., Khan, M. S., Dhruba, A. R., Khan, M. M., Al-Amri, J. F., Masud, M., & Rawashdeh, M. (2021). Deep learning-based sentiment analysis of COVID-19 vaccination responses from Twitter data. Computational and Mathematical Methods in Medicine, 2021.
6. Hu, N. (2022). Sentiment analysis of texts on public health emergencies based on social media data mining. Computational and mathematical methods in medicine, 2022.
7. Edara, D. C., Vanukuri, L. P., Sistla, V., & Kolli, V. K. K. (2023). Sentiment analysis and text categorization of cancer medical records with LSTM. Journal of Ambient Intelligence and Humanized Computing, 14(5), 5309-5325.
8. Yuvaraj, R., Thagavel, P., Thomas, J., Fogarty, J., & Ali, F. (2023). Comprehensive analysis of feature extraction methods for emotion recognition from multichannel EEG recordings. *Sensors*, *23*(2), 915.
9. Wang, F., Ding, Z., Xia, R., Li, Z., & Yu, J. (2022). Multimodal emotion-cause pair extraction in conversations. IEEE Transactions on Affective Computing.
10. Denecke, K., & Reichenpfader, D. (2023). Sentiment analysis of clinical narratives: a scoping review. *Journal of Biomedical Informatics*, 104336.

11. Food, U. S. (2019). Manufacturer and user facility device experience. http://www. accessdata. fda. gov/scripts/cdrh/cfdocs/cfmaude/search. cfm.
12. Bala, I., Kelly, T. L., Stanford, T., Gillam, M. H., & Mitchell, L. (2024). Machine learning-based analysis of adverse events in mesh implant surgery reports. Social Network Analysis and Mining, 14(1), 63.
13. Mangir, N., Roman, S., Chapple, C. R., & MacNeil, S. (2020). Complications related to use of mesh implants in surgical treatment of stress urinary incontinence and pelvic organ prolapse: infection or inflammation?. *World Journal of Urology*, *38*, 73-80.
14. Hollmer, M. (2010, June 5). J&J/Ethicon to stop selling four vaginal mesh implants. Retrieved from https://www.fiercebiotech.com/medical-devices/j-j-ethicon-to-stop-selling-four-vaginal-mesh-implants
15. D. Voreacos, A. Nussbaum, "The next medical device controversy: vaginal mesh", Business Week. 2011.
16. T. Watanabe T, MB. Chancellor, "Pelvic surgeons caught in the meshes of the law", Rev Urol. Vol. 14(1-2), PP. 35-36, 2012.
17. Dyer, C. (2020). Mesh implants: Women launch claims against NHS trusts and surgeons for failing to warn of risks. BMJ: British Medical Journal (Online), 369.
18. U.S. Food and Drug Administration. (2019, April 16). FDA takes action to protect women's health, orders manufacturers of surgical mesh intended for transvaginal repair of pelvic organ prolapse to stop selling all devices. Retrieved from https://www.fda.gov/news-events/press-announcements/fda-takes-action-protect-womens-health-orders-manufacturers-surgical-mesh-intended-transvaginal
19. Ng-Stollmann, N., Fünfgeld, C., Gabriel, B., & Niesel, A. (2020). The international discussion and the new regulations concerning transvaginal mesh implants in pelvic organ prolapse surgery. *International urogynecology journal*, *31*(10), 1997-2002.
20. Farmer, Z. L., Utomo, E., Domínguez-Robles, J., Mancinelli, C., Mathew, E., Larrañeta, E., & Lamprou, D. A. (2021). 3D printed estradiol-eluting urogynecological mesh implants: Influence of material and mesh geometry on their mechanical properties. *International Journal of Pharmaceutics*, *593*, 120145.
21. Ensign, L. G., & Cohen, K. B. (2017). A primer to the structure, content and linkage of the FDA's Manufacturer and User Facility Device Experience (MAUDE) files. eGEMs, 5(1).
22. Wang, M., & Hu, F. (2021). The application of nltk library for python natural language processing in corpus research. Theory and Practice in Language Studies, 11(9), 1041-1049.
23. Bala, I., Kelly, T. L., Lim, R., Gillam, M. H., & Mitchell, L. (2022). An Effective Approach for Multiclass Classification of Adverse Events Using Machine Learning. Journal of Computational and Cognitive Engineering.
24. Loria, S. (2018). textblob Documentation. Release 0.15, 2(8), 269.
25. Abiola, O., Abayomi-Alli, A., Tale, O. A., Misra, S., & Abayomi-Alli, O. (2023). Sentiment analysis of COVID-19 tweets from selected hashtags in Nigeria using VADER and Text Blob analyser. Journal of Electrical Systems and Information Technology, 10(1), 5.
26. Mohammad, S. M., & Turney, P. D. (2013). Nrc emotion lexicon. National Research Council, Canada, 2, 234.
27. Khawaja, H. S., Beg, M. O., & Qamar, S. (2018, November). Domain specific emotion lexicon expansion. In 2018 14th international conference on emerging technologies (ICET) (pp. 1-5). IEEE.
28. Le, T. N., Afshar Ali, M., Gadzhanova, S., Lim, R., Bala, I., Bogale, K. N., & Gillam, M. (2024). Hernia repair prevalence by age and gender among the Australian adult population from 2017 to 2021. *Critical Public Health, 34*(1), 1-11.